\documentclass[10pt,twocolumn,letterpaper]{article}

\usepackage{cvpr}
\usepackage{times}
\usepackage{epsfig}
\usepackage{graphicx}
\usepackage{amsmath}
\usepackage{amssymb}

\usepackage{multirow}
\usepackage{booktabs}
\usepackage{mathtools}  
\usepackage{nicefrac}

\usepackage[pagebackref=true,breaklinks=true,letterpaper=true,colorlinks,bookmarks=false]{hyperref}

\begin{document}
	
	\newcommand{\x}{\boldsymbol{x}}
	\newcommand{\ttheta}{\boldsymbol{\theta}}
	\newcommand{\ddelta}{\boldsymbol{\delta}}
	\newcommand{\w}{\boldsymbol{w}}
	\newcommand{\len}{\boldsymbol{l}}
	\newcommand{\M}{\boldsymbol{M}}
	\newcommand{\R}{\mathbb{R}}
	\newcommand{\J}{\mathcal{J}}

	\newtheorem{definition}{Definition}
	
	\title{Immune Defense: A Novel Adversarial Defense Mechanism for Preventing the Generation of Adversarial Examples}
	
	\author{Jinwei Wang, Hao Wu, Haihua Wang, Jiawei Zhang\\
	Nanjing University of Information Science and Technology
	\and
	Xiangyang Luo\\
	State Key Laboratory of Mathematical Engineering and Advanced Computing
	\and
	Bin Ma \\
	Qilu University of Technology
}

\maketitle

\begin{abstract}
	\small
	The vulnerability of Deep Neural Networks (DNNs) to adversarial examples has been confirmed. Existing adversarial defenses primarily aim at preventing adversarial examples from attacking DNNs successfully, rather than preventing their generation. If the generation of adversarial examples is unregulated, images within reach are no longer secure and pose a threat to non-robust DNNs. Although gradient obfuscation attempts to address this issue, it has been shown to be circumventable. Therefore, we propose a novel adversarial defense mechanism, which is referred to as immune defense and is the example-based pre-defense. This mechanism applies carefully designed quasi-imperceptible perturbations to the raw images to prevent the generation of adversarial examples for the raw images, and thereby protecting both images and DNNs. These perturbed images are referred to as Immune Examples (IEs). In the white-box immune defense, we provide a gradient-based and an optimization-based approach, respectively. Additionally, the more complex black-box immune defense is taken into consideration. We propose Masked Gradient Sign Descent (MGSD) to reduce approximation error and stabilize the update to improve the transferability of IEs and thereby ensure their effectiveness against black-box adversarial attacks. The experimental results demonstrate that the optimization-based approach has superior performance and better visual quality in white-box immune defense. In contrast, the gradient-based approach has stronger transferability and the proposed MGSD significantly improve the transferability of baselines.
\end{abstract}

\section{Introduction}

\begin{figure}[t]
	\begin{center}
		\begin{minipage}[b]{0.48\linewidth}
			\begin{center}
				\includegraphics[width=1.0\linewidth]{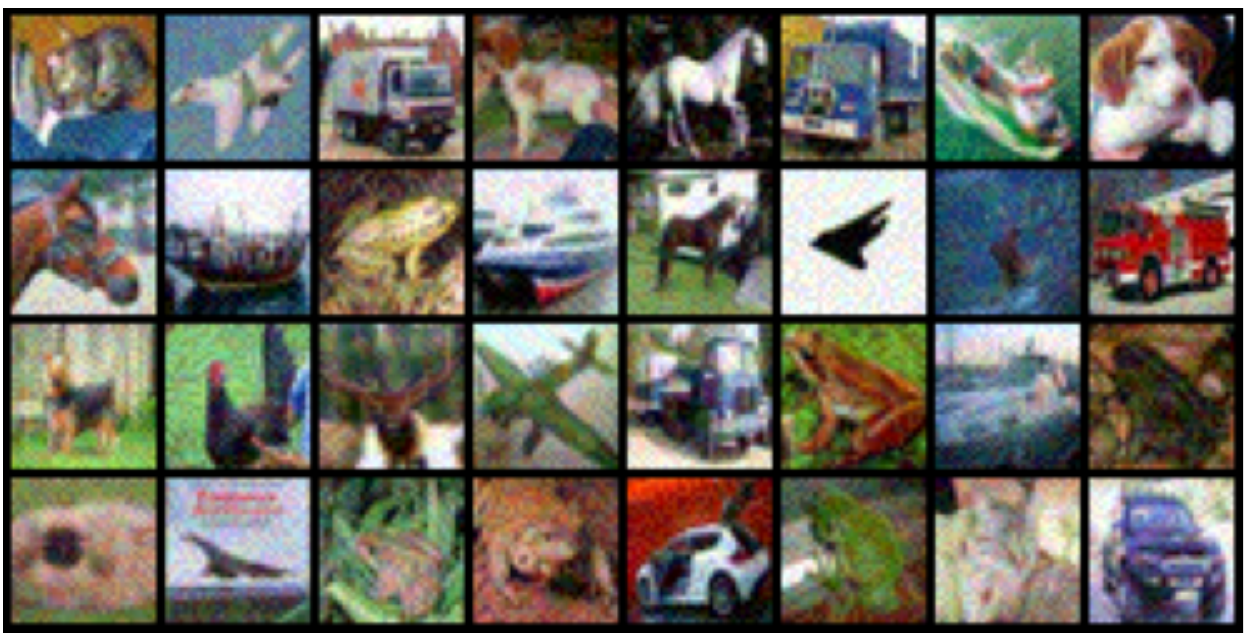}
			\end{center}
		\end{minipage}
		\begin{minipage}[b]{0.48\linewidth}
			\begin{center}
				\includegraphics[width=1.0\linewidth]{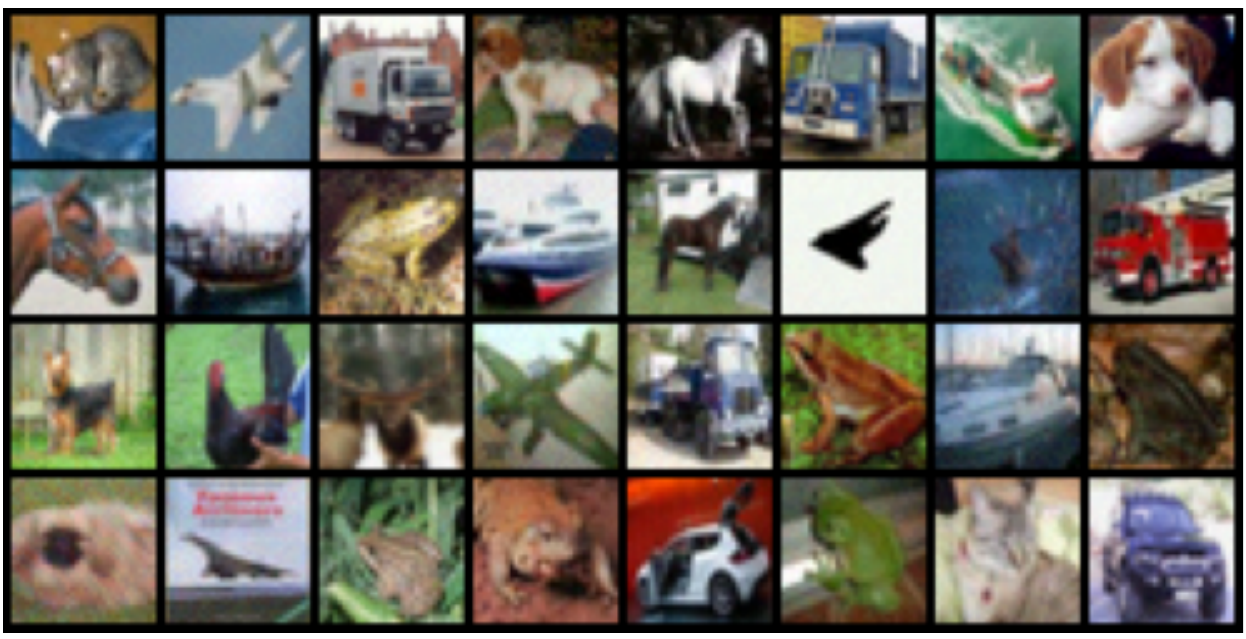}
			\end{center}
		\end{minipage}
		\begin{minipage}[b]{0.48\linewidth}
			\begin{center}
				\includegraphics[width=1.0\linewidth]{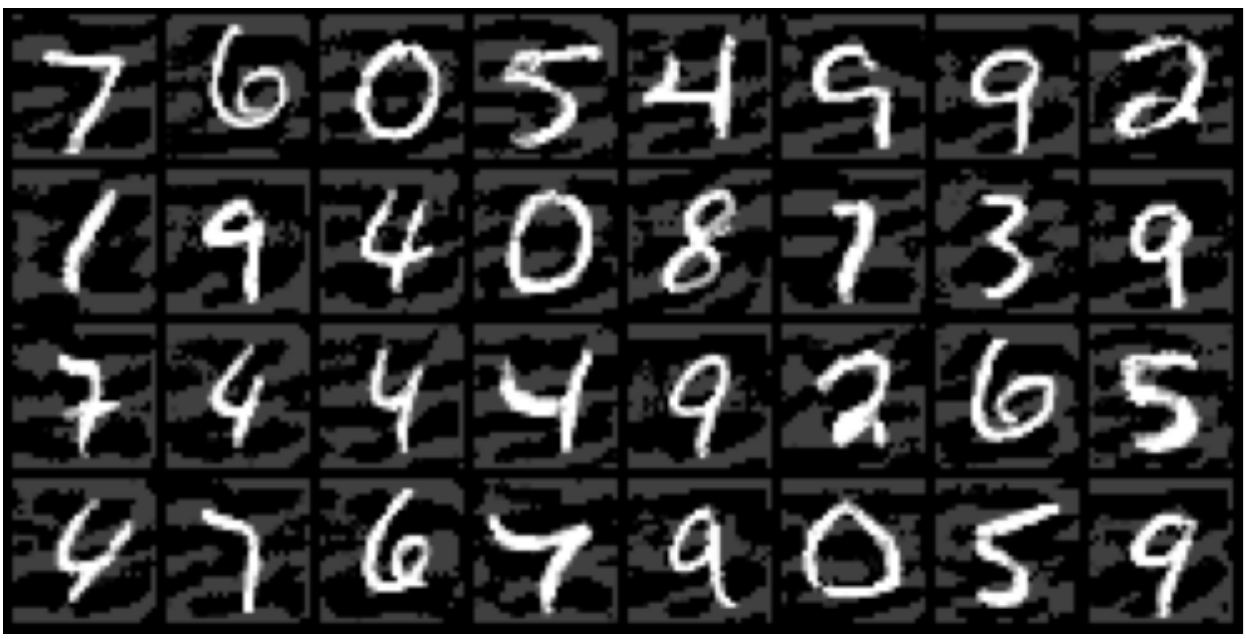}
			\end{center}
		\end{minipage}
		\begin{minipage}[b]{0.48\linewidth}
			\begin{center}
				\includegraphics[width=1.0\linewidth]{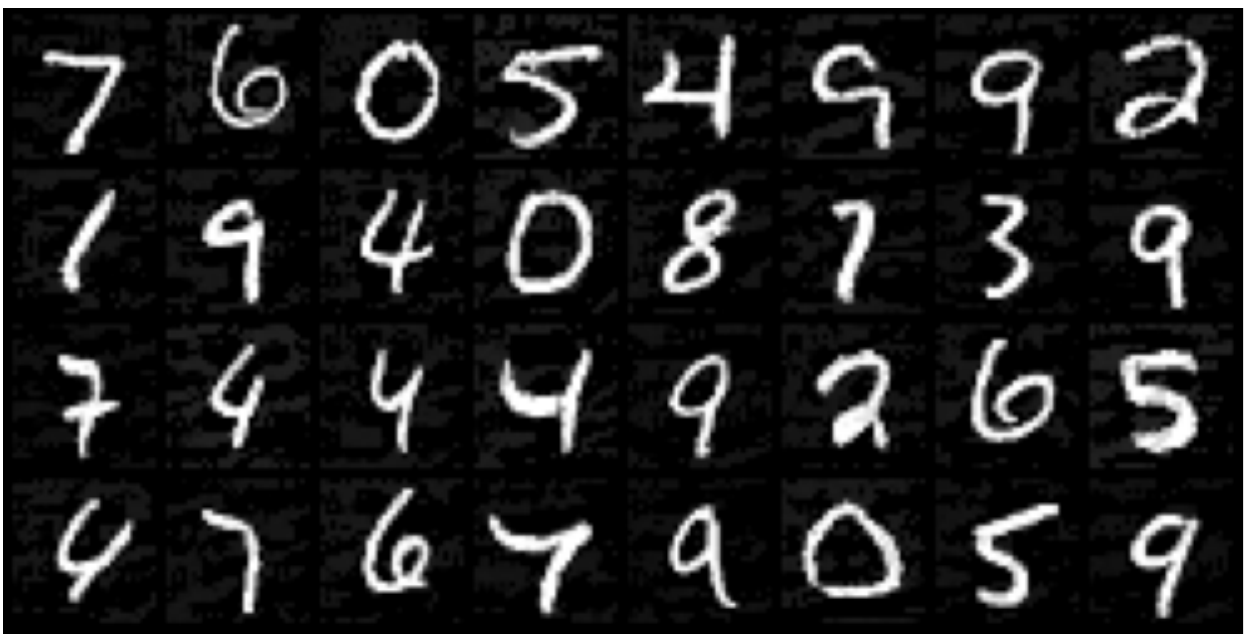}
			\end{center}
		\end{minipage}
	\end{center}
	\caption{White-box immune examples crafted for AdvGAN on CIFAR-10 and MNIST, respectively. Top: CIFAR-10. Bottom: MNIST. Left: Grad. (gradient-based approach). Right: Opt. (optimization-based approach).}
	\label{fig:wbie}
\end{figure}
Adversarial examples~\cite{szegedy2013intriguing,goodfellow2014explaining,papernot2016transferability,huang2017adversarial,papernot2017practical,tramer2017space} have demonstrated the inherent vulnerability of Deep Neural Networks (DNNs)~\cite{szegedy2016rethinking,szegedy2017inception,he2016deep,simonyan2014very}. Attackers can easily craft such examples by applying carefully designed and quasi-imperceptible perturbations into clean examples. When adversarial examples are fed into DNNs, they can cause networks to produce incorrect outputs, leading to serious security concerns.

On the one hand, adversarial attacks can be classified into three main types: gradient-based attacks~\cite{goodfellow2014explaining,kurakin2016adversarial,dong2018boosting,lin2019nesterov,wang2021enhancing,madry2017towards}, optimization-based attacks~\cite{szegedy2013intriguing,carlini2017towards}, and generation-based attacks~\cite{baluja2018learning,hayes2018learning,xiaogenerating,jandial2019advgan++,zhang2022self}.
Specifically, gradient-based attacks utilize the gradient of the DNN to increase the classification loss for crafting adversarial examples, which have the advantages of fast attack speed and high transferability, but their attack capability is limited. In contrast, optimization-based attacks optimize a multi-objective loss function to craft strong adversarial examples with high visual quality. Nevertheless, they have slow attack speed and poor transferability. Furthermore, generation-based attacks can directly generate adversarial examples by deep generative models~\cite{masci2011stacked,goodfellow2020generative,radford2015unsupervised} without accessing the DNN, which provides both fast generation speed and high attack capability.

On the other hand, popular adversarial defenses include adversarial training~\cite{goodfellow2014explaining,tramer2017ensemble,madrytowards}, defensive distillation~\cite{papernot2016distillation}, input pre-processing~\cite{xiemitigating,xu2017feature,guocountering,liu2019feature}, detection-based methods~\cite{feinman2017detecting,carlini2017adversarial}, and gradient obfuscation~\cite{guocountering,xiemitigating,songpixeldefend,athalye2018obfuscated}. These methods defend against adversarial attacks by improving model robustness, detecting adversarial examples, or feeding false gradients to the attacker. Except for gradient obfuscation, all adversarial defenses are post-defenses against adversarial examples. They aim to prevent adversarial examples from attacking DNNs successfully but do not help to avoid the generation of adversarial examples. Without controlling the generation of adversarial examples, the images within reach are no longer secure and could pose a threat to non-robust DNNs at any time. Gradient obfuscation appears to increase the difficulty of generating adversarial examples, but Athalye~\etal~\cite{athalye2018obfuscated} pointed out that it can be easily circumvented and thereby gives a false sense of security. Generally speaking, existing adversarial defenses cannot restrict the generation of adversarial examples and thereby are unable to protect images.

In response to this issue, we propose a novel adversarial defense mechanism, \ie, immune defense, that involves applying carefully designed perturbations to raw images to prevent the generation of adversarial examples. We refer to such perturbed images as Immune Examples (IEs), and the applied perturbations as Immune Perturbations (IPs). This mechanism enhances the security of both images and DNNs, mitigating the threat of potential attacks. The proposed immune defense is required not only to disable the adversarial attacks but also to ensure the correct classification of images. This dual requirement makes the task of immune defense more challenging. To the best of our knowledge, no prior research has investigated the immune defense in a similar manner.

In the white-box immune defense, where the defenders know complete knowledge of the target adversarial attack, they can easily craft IEs with strong performance to prevent the generation of adversarial examples. In this work, we give a gradient-based and an optimization-based approach, respectively, for white-box immune defense and compare their performance in Sec.~\ref{sec:wid}. Fig.~\ref{fig:wbie} displays the IEs generated by such two approaches.

In addition, the more complex black-box immune defense should also be taken into account. We need to consider the potential threat that defenders craft white-box IEs against a specific adversarial attack, but these IEs are rendered ineffective against other black-box adversarial attacks, which means that unkown attackers can generate effective adversarial examples for these IEs. To address this issue, we propose the Masked Gradient Sign Descent (MGSD) to improve the transferability of IEs, making IEs prevent various adversarial attacks from generating adversarial examples, which will be discussed in Sec.~\ref{sec:gmm}. The proposed method masks the gradients to reduce approximation error and stabilize the update. Extensive experiments in the black-box immune defense demonstrate that the proposed method significantly improves the transferability of IEs across various adversarial attacks. 

In conclusion, the contributions are summarized as follows:
\begin{itemize}
	\item We propose a novel adversarial defense mechanism, \ie, immune defense, which is the example-based pre-defense, rather than model-based defense or post-defense. The immune defense prevents the generation of adversarial examples to protect the images from damage by adversarial perturbations.
	\item We provide two white-box immune defense approaches, \ie, gradient-based and optimization-based approach, and propose the Masked Gradient Sign Descent (MGSD) for black-box immune defense to reduce approximation error and stabilize the update to improve the transferability of IEs.
	\item Extensive experiments demonstrate that the white-box immune defense can generate IEs with strong performance and high visual quality. Furthermore, for black-box defense, the proposed MGSD is superior to the baselines in terms of transferring IEs to other adversarial attacks.
\end{itemize}
\begin{figure*}[t]
	\begin{center}
		\includegraphics[width=0.95\linewidth]{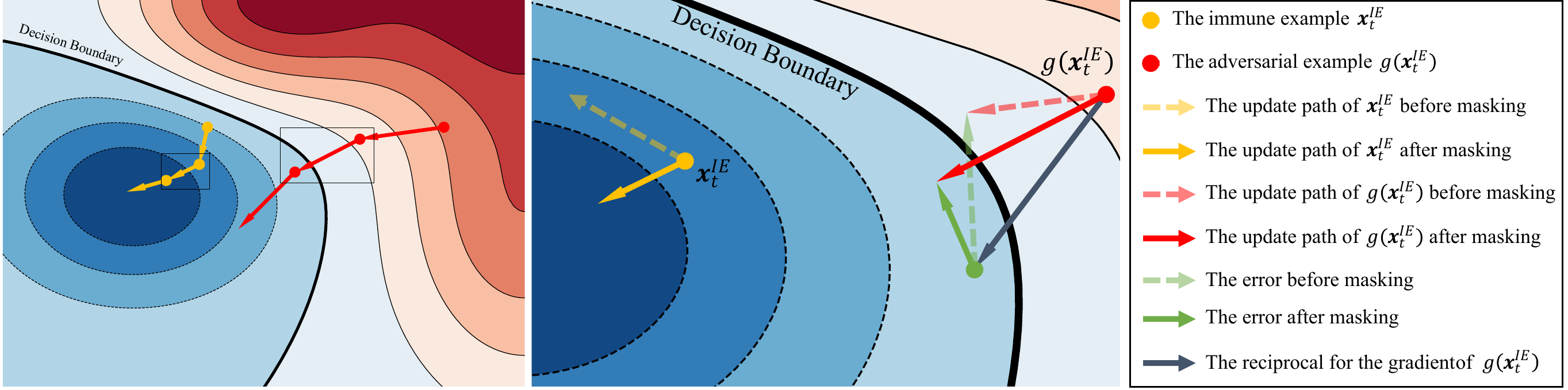}
	\end{center}
	\caption{Illustration of MGSD. MGSD reduces errors through masking, accelerating the optimization of adversarial examples to the decision region of the ground-truth label and stabilizing the update of IEs, thus improving the transferability of IEs.}
	\label{fig:mgsd_ill}
\end{figure*}

\section{Background}

Since generation-based attacks can fastly generate diverse adversarial examples with strong performance and high visual quality, they have become an important target of our defense in this work. In this section, we introduce several optimization-based attacks and related works on the transferability of adversarial examples.

\subsection{Generation-based attacks}

Generation-based attacks utilize deep generative models to generate adversarial examples. Unlike gradient-based and optimization-based attacks, generation-based attacks train deep generative models on the target classifier and the original data to generate adversarial examples that closely resemble the original data. Once the deep generative model has been trained, the attacker can efficiently generate diverse adversarial examples without the target classifier. For example, the Adversarial Transformation Network (ATN)~\cite{baluja2018learning} with autoencoder architecture generates targeted adversarial examples that minimally perturb both the original input and the output of the target classifier. Furthermore, the Universal Adversarial Network (UAN)~\cite{hayes2018learning} generates Universal Adversarial Perturbations (UAPs)~\cite{moosavi2017universal} by decoding random noise, which can be applied to any example from the dataset. Apart from autoencoder-based and decoder-based attacks, AdvGAN~\cite{xiaogenerating} utilizes the Generative Adversarial Network (GAN)~\cite{goodfellow2020generative,radford2015unsupervised} to generate high perceptual quality and more effective adversarial examples. On the basis of AdvGAN, AdvGAN++~\cite{jandial2019advgan++} replaces the original encoder with the target model to obtain more vulnerable latent features, resulting in stronger adversarial examples. Regarding privacy protection, Recoverable GAN (RGAN)~\cite{zhang2022self} has shown to be an effective solution. The generator of RGAN applies adversarial perturbations to images to prevent malicious detection and analysis by intelligent algorithms. Additionally, its recovery component removes the adversarial perturbations nearly losslessly, recovering the visual quality of images.

\subsection{Transfer-based attacks}

Gradient-based attacks have been demonstrated to be effective in improving the transferability of adversarial examples. Momentum Iterative Fast Gradient Sign Method (MI-FGSM)~\cite{dong2018boosting} integrated Polyak momentum~\cite{polyak1964some} into Iterative Fast Gradient Sign Method (I-FGSM)~\cite{kurakin2016adversarial} to stabilize updates and help adversarial examples escape from poor localmaxima to improve transferability. Inspired by the fact that Nesterov momentum~\cite{Nesterov1983AMF} is superior to Polyak momentum, Nesterov I-FGSM (NI-FGSM)~\cite{lin2019nesterov} integrated Nesterov momentum into I-FGSM to further improve transferability. Different from previous momentum-based attacks, Variance tuning MI-FGSM (VMI-FGSM)~\cite{wang2021enhancing} adopted the gradient variance of the previous iteration to tune the current gradient. The idea is to reduce the variance of the gradient at each iteration to stabilize the update and escape from saddle points and poor local extrema.

\section{Immune defense}

In this section, we first define the research content of this work. We then provide a detailed description of the proposed white box and black box immune defense methods, respectively.

\subsection{Problem definition}

Adversarial attacks cause DNNs to output incorrect predictions (\ie, non-targeted attacks), or even specified incorrect classes (\ie, targeted attacks). However, the immune defense can disable adversarial attacks from generating adversarial examples for IEs by applying immune perturbations to raw images. Specifically, we define the IE as follows.
\begin{definition}
	Given a raw image $\x \in \R^n$ with a ground-truth label $y \in \{1, \cdots, l\}$, a classifier $f: \R^n \to \{1, \cdots, l\}$ with a loss function $\J(\x, y): \R^n \times \{1, \cdots, l\} \to \R^+$ (\eg, cross-entropy), and an adversarial attack algorithm $g: \x \mapsto g(\x)$ that inputs the raw image $\x$ and outputs a corresponding adversarial example $g(\x) \in \R^n$, $\x^{IE}$ is an Immune Example (IE) of the raw image $\x$, which satisfies
	\begin{equation}
		\begin{gathered}
			\begin{cases}
				f(\x^{IE}) = y, \\
				f(g(\x^{IE})) = y,
			\end{cases} \\
			\mathrm{s.t.}~\left \| \x^{IE} - \x \right \|_p \le \tau,
		\end{gathered}
		\label{eq:ie_definition}
	\end{equation}
	where $\left \| \cdot \right \|_p$ denotes the $L_p$ norm, $\x^{IE} - \x$ is referred to as the Immune Perturbation (IP), and $\tau$ denotes the size of immune perturbation used to regulate the imperceptibility of the IP.
	\label{def:ie}
\end{definition}

The adversarial attack $g$ may be a white-box or a black-box, and the immune defense against the white-box (or black-box) adversarial attack is referred to as white-box (or black-box) immune defense. To ease the study, we make the following basic assumptions:
\begin{enumerate}
	\item \label{list:one} Assuming that the attacker has complete knowledge of the target classifier and conducts a white-box adversarial attack. The attacker can easily generate strong adversarial examples, and the IEs must remain effective in this worst case.
	\item \label{list:two} Assuming that the defender also has complete knowledge of the target classifier. This assumption is based on the fact that the defender and the model publisher face common adversaries. The model publisher can further defend against adversarial attacks by crafting IEs.
\end{enumerate}

\subsection{White-box immune defenses}\label{sec:whitebox}

In the white-box immune defense, the complete knowledge of the adversarial attack $g$ is accessible. According to Definition~\ref{def:ie} and Assumption~\ref{list:two}, we can obtain the IE simply by minimizing the classification loss of the IE and the corresponding adversarial example as follows:
\begin{equation}
	\begin{gathered}
		\underset{\x^{IE}}{\arg \min}~\lambda \cdot \J(\x^{IE}, y) + \J(g(\x^{IE}), y), \\
		\mathrm{s.t.}~\left \| \x^{IE} - \x \right \|_p \le \tau,
	\end{gathered}
	\label{eq:ie_optim}
\end{equation}
where $\J(\x^{IE}, y)$ is referred to as the immune classification loss, which represents the classification loss of the IE, $\J(g(\x^{IE}), y)$ is referred to as the adversarial classification loss, which represents the classification loss of the corresponding adversarial example, and $\lambda$ denotes the weight of the immune classification loss. Eq.~\ref{eq:ie_optim} is a gradient-based approach, and inspired by gradient-based attacks~\cite{goodfellow2014explaining,kurakin2016adversarial}, we solve it by vanilla Gradient Sign Descent (GSD)~\cite{riedmiller1993direct,goodfellow2014explaining,kurakin2016adversarial,papernot2016transferability} under the constraint of $L_{\infty}$ norm as follows:
\begin{equation}
	\nabla_{\x_t^{IE}} = \lambda \cdot \nabla_{\x_t^{IE}} \J(\x_t^{IE}, y) + \nabla_{\x_t^{IE}} \J(g(\x_t^{IE}), y),
	\label{eq:ensemble_grad}
\end{equation}
\begin{equation}
	\x_{t+1}^{IE} = \mathrm{Clip}_{(\x, \tau)} \{ \x_t^{IE} - \alpha \cdot \mathrm{sign}(\nabla_{\x_t^{IE}}) \},
	\label{eq:ie_update}
\end{equation}
where $\x_t^{IE}$ denotes the IE in the $t$-th iteration, $\mathrm{Clip}_{(\x, \tau)} \{ \cdot \}$ restricts the input to be within the $\tau$-ball of $\x$, $\alpha$ denotes the step size, and $\mathrm{sign}(\cdot)$ denotes the sign function.

Additionally, to deal with the box-constrained IP, we also use the following optimization-based approach to approximate Eq.~\ref{eq:ie_optim}:
\begin{equation}
	\begin{gathered}
		\underset{\x^{IE}}{\arg \min}~\!\lambda\!\cdot\!\J\!(\x^{IE}\!,\!y)\!+\!\J(g(\x^{IE})\!,\!y)\!+\!\eta\!\cdot\!\left \| \x^{IE}\!-\!\x \right \|_p\!,\!
	\end{gathered}
	\label{eq:ie_opt1}
\end{equation}
where $\eta$ denotes the weight of the IP loss. Inspired by optimization-based attacks~\cite{carlini2017towards,szegedy2013intriguing}, we use Adam~\cite{kingma2014adam} to solve Eq.~\ref{eq:ie_opt1} and compare the performance of such two white-box immune defense approaches in Sec.~\ref{sec:wid}.

\subsection{Black-box immune defenses}\label{sec:gmm}

In the black-box setting, the defender can use black-box optimization algorithms~\cite{holland1992genetic,kennedy1995particle,storn1996usage,chen2017zoo} to directly solve the optimization problem introduced in Sec.~\ref{sec:whitebox}. However, this idea is straightforward and falls outside the scope of this work. Instead, we consider a more common and challenging case where IEs are fed into black-box adversarial attacks other than the white-box adversarial attack. In this case, IEs must be transferable to remain effective. Therefore, we focus on transfer-based black-box immune defense in this work. 

Since minimizing the adversarial classification loss can return adversarial examples into the decision region of the ground-truth label, and the decision boundaries of different DNNs are similar~\cite{liu2016delving}, it is possible to transfer IEs across various adversarial attacks. Generally, the lower adversarial classification loss indicates a higher likelihood of returning the adversarial examples into the decision region of the ground-truth label. However, extremely low adversarial classification loss also raises the risk of overfitting IEs to the source adversarial attack, which reduces the generalizability of IEs to other adversarial attacks. Therefore, only by decreasing the adversarial classification loss to an appropriate degree can the transferability of IEs be effectively improved.

In addition, gradient-based adversarial attacks have higher transferability than optimization-based adversarial attacks but lower white-box attack capability~\cite{papernot2016transferability}, because optimization-based adversarial attacks overfit the adversarial examples to the source model more than gradient-based adversarial attacks~\cite{kurakinadversarial,dong2018boosting}. Inspired by this fact, we reasonably hypothesize that gradient-based immune defenses also have higher transferability than optimization-based immune defenses but lower performance in the white-box immune defense, and we validate this hypothesis in Sec.~\ref{sec:wid}. Therefore, we are more concerned with gradient-based immune defenses to avoid extreme overfitting of IEs and sufficiently decrease the adversarial classification loss to ensure the transferability of IEs in this work.

Next, we attempt to explore how the gradient-based approach decreases the adversarial classification loss. In the subsequent analysis, we omit the immune classification loss and only focus on the adversarial classification loss for simplicity. Specifically, we update the IE according to Eq.~\ref{eq:ensemble_grad} and~\ref{eq:ie_update}, and set $\lambda$ to $0$:
\begin{equation}
	\len = - \alpha \cdot \mathrm{sign}(\nabla_{\x_t^{IE}} \J(g(\x_t^{IE}), y)),
\end{equation}
\begin{equation}
	\x_{t+1}^{IE} = \mathrm{Clip}_{(\x, \tau)} \{ \x_t^{IE} + \len \}.
	\label{eq:ie_update2}
\end{equation}

Then, the variation $\Delta_k^t$ in the $k$-th dimension for the corresponding adversarial examples of $\x_t^{IE}$ and $\x_{t+1}^{IE}$ is calculated as follows\footnote{The proofs and analysis of some formulas in this subsection are written in the Appendix.}:
\begin{align}
	\Delta_k^t
	&= g(\x_{t+1}^{IE})_k - g(\x_t^{IE})_k \notag\\
	&= \frac{g(\x_t^{IE} + \len)_k - g(\x_t^{IE})_k}{\left \| \len \right \|_2} \cdot \left \| \len \right \|_2 \notag\\
	&= \left( \nabla_{\len} g(\x_t^{IE})_k + e(\left \| \len \right \|_2) \right) \cdot \left \| \len \right \|_2 \notag\\
	&=-\alpha\!\cdot\!\frac{\left \| \nabla_{\x_t^{IE}} \J(g(\x_t^{IE}), y) \right \|_1}{\nabla_{g(\x_t^{IE})_k} \J(g(\x_t^{IE}), y)}\!+\!\left \| \len \right \|_2\!\cdot\!e(\left \| \len \right \|_2),
	\label{eq:adv_delta}
\end{align}
where $g(\x_t^{IE})_k$ denotes the $k$-th dimension of the adversarial example $g(\x_t^{IE})$, $e(\left \| \len \right \|_2) \sim N(0, \sigma \cdot \left \| \len \right \|_2)$ denotes the approximation error of the directional derivative $\nabla_{\len} g(\x_t^{IE})_k$, $\sigma > 0$ denotes the standard deviation factor. The approximation error is introduced due to not satisfying $\left \| \len \right \|_2 \to 0$. To account for the fact that
\begin{equation}
	\lim_{\left \| \len \right \|_2 \to 0} e(\left \| \len \right \|_2) \to 0,\label{eq:error_limit}
\end{equation}
we assume that the approximation error roughly follows a normal distribution with mean $0$ and variance positively correlated with $\left \| \len \right \|_2$.

It can be observed from Eq.~\ref{eq:adv_delta} that if the IE is updated according to Eq.~\ref{eq:ie_update2}, then the corresponding adversarial example is updated according to as follows:
\begin{equation}
	\resizebox{.88\hsize}{!}{$
		g(\x_{t+1}^{IE}) = g(\x_t^{IE}) -\alpha \cdot \frac{\left \| \nabla_{\x_t^{IE}} \J(g(\x_t^{IE}), y) \right \|_1}{\nabla_{g(\x_t^{IE})} \J(g(\x_t^{IE}), y)} + \left \| \len \right \|_2 \cdot e(\left \| \len \right \|_2)^n.
		$}
	\label{eq:adv_update}
\end{equation}

Intuitively, the adversarial example is updated approximately by the gradient reciprocal descent method with a step size $\alpha \cdot \left \| \nabla_{\x_t^{IE}} \J(g(\x_t^{IE}), y) \right \|_1$, and the approximation error is represented by the error term $\left \| \len \right \|_2 \cdot e(\left \| \len \right \|_2)^n$. Although the reciprocal of the gradient is not superior to the gradient for optimization, the sign of the gradient reciprocal is consistent with that of the gradient, which enables the gradient reciprocal descent method to also decrease the adversarial classification loss. However, in the dimension where the sign of the error is consistent with that of the gradient, the error inhibits the decrease of the adversarial classification loss. Therefore, reducing the error can speed up the decrease of the adversarial classification loss and consequently improve the transferability of the IE.

Naturally, we next analyze the error term $\left \| \len \right \|_2 \cdot e(\left \| \len \right \|_2)^n$. We rewrite $\left \| \len \right \|_2$ as follows:
\begin{align}
	\left \| \len \right \|_2 = \alpha \cdot \sqrt{\left \| \nabla_{\x_t^{IE}} \J(g(\x_t^{IE}), y) \right \|_0}.
\end{align}
This equation shows that the error decreases as $\left \| \nabla_{\x_t^{IE}} \J(g(\x_t^{IE}), y) \right \|_0$ decreases. Intuitively, properly masking $\nabla_{\x_t^{IE}} \J(g(\x_t^{IE}), y)$ can lead to the reduction of the error. Specifically, we adopt the following mask $\M$:
\begin{align}
	\M_k = 
	\begin{cases}
		1 & \text{if}~\frac{\nabla_{{\x_t^{IE}}_k} \J(g(\x_t^{IE}), y)}{\nabla_{{g(\x_t^{IE})}_k} \J(g(\x_t^{IE}), y)}  > 0, \\
		0 & \text{otherwise},
	\end{cases}
	\label{eq:mask}
\end{align}
where $\M_k$ and ${\x_t^{IE}}_k$ denote the $k$-th dimension of $\M$ and $\x_t^{IE}$, respectively. This mask eliminates the dimensions where the update direction of $\x_t^{IE}$ doesn't align with that of $g(\x_t^{IE})$ and only keeps the aligned dimensions, which prevents $\x_t^{IE}$ from oscillating in some dimensions and thereby stabilizes the update. In brief, we can both reduce the error and stabilize the update of the IE by such a mask, thereby improving the transferability of the IE. The illustration of this method is shown in Fig.~\ref{fig:mgsd_ill}.

Finally, we directly take the immune classification loss into consideration and rewrite Eq.~\ref{eq:ensemble_grad} as follows:
\begin{equation}
	\resizebox{.85\hsize}{!}{$
		\nabla_{\x_t^{IE}} = \lambda \cdot \nabla_{\x_t^{IE}} \J(\x_t^{IE}, y) + \M \odot \nabla_{\x_t^{IE}} \J(g(\x_t^{IE}; \ttheta_g), y),
		$}
	\label{eq:masked_ensemble_grad}
\end{equation}
where $\odot$ denotes the Hadamard product, and $\M$ is given by Eq.~\ref{eq:mask}. We refer to the proposed method, \ie Eq.~\ref{eq:masked_ensemble_grad} and Eq.~\ref{eq:ie_update}, as the Masked Gradient Sign Descent (MGSD).

\subsection{Immune rate}

\begin{table}[t]
	\begin{minipage}[b]{.48\linewidth}
		\begin{center}
			\begin{tabular}{|lc|cc|}
				\hline
				\multicolumn{2}{|l|}{\multirow{2}{*}{Results}}  & \multicolumn{2}{c|}{$A^{adv}$}    \\ \cline{3-4} 
				\multicolumn{2}{|l|}{}                       & \multicolumn{1}{c|}{T} & F \\ \hline
				\multicolumn{1}{|l|}{\multirow{2}{*}{$A$}} & T & \multicolumn{1}{c|}{$A_1$} & $A_2$ \\ \cline{2-4} 
				\multicolumn{1}{|l|}{}                   & F & \multicolumn{1}{c|}{$A_3$} & $A_4$ \\ \hline
			\end{tabular}
		\end{center}
		\caption{The confusion matrix of the classification results for $\x$ and $g(\x)$.}
		\label{tab:cf1}
	\end{minipage}
	\hfill
	\begin{minipage}[b]{.48\linewidth}
		\begin{center}
			\begin{tabular}{|lc|cc|}
				\hline
				\multicolumn{2}{|l|}{\multirow{2}{*}{Results}}  & \multicolumn{2}{c|}{$B^{adv}$}    \\ \cline{3-4} 
				\multicolumn{2}{|l|}{}                       & \multicolumn{1}{c|}{T} & F \\ \hline
				\multicolumn{1}{|l|}{\multirow{2}{*}{$B$}} & T & \multicolumn{1}{c|}{$B_1$} & $B_2$ \\ \cline{2-4} 
				\multicolumn{1}{|l|}{}                   & F & \multicolumn{1}{c|}{$B_3$} & $B_4$ \\ 
				\hline
			\end{tabular}
		\end{center}
		\caption{The confusion matrix of the classification results for $\x^{IE}$ and $g(\x^{IE})$.}
		\label{tab:cf2}
	\end{minipage}
\end{table}
In this section, we propose the Immune Rate (IR) to measure the performance of the IEs. To aid in the description, we first provide the confusion matrix of classification results for the raw image set $A$ and the corresponding adversarial example set $A^{adv} = \{ g(\x) | \x \in A \}$, as shown in Table~\ref{tab:cf1}. We use the notation ``T" to represent correct classification, ``F" to represent misclassification, and $A_{1-4}$ to represent the set of $\x$ for various classification results, respectively. Then, the Attack Success Rate (ASR) of $A^{adv}$ against $f$ is defined as
\begin{equation}
	ASR(A^{adv}; f) = \frac{\left | A_2 \right |}{\left | A_1 \cup A_2 \right |},
	\label{eq:asr}
\end{equation}
where $\left | \cdot \right |$ denotes the cardinal number of a given set. It is easy to observe from Eq.~\ref{eq:asr} that the raw images that are misclassified are ignored because they are highly likely to be successfully crafted as effective adversarial examples, and such examples cannot comprehensively reflect the performance of adversarial examples. Similarly, $ASR(B^{adv}; f)$ can be defined according to Table~\ref{tab:cf2}, where $B^{adv} = \{ g(\x^{IE}) | \x^{IE} \in B \}$, $\x^{IE}$ denotes the IE of $\x$, $B$ denotes the set of IEs, $B_{1-4}$ denotes the set of $\x$ for various classification results, respectively.

The performance of IEs can be simply measured by the Variation of the ASR (VASR), \ie,
\begin{equation}
	\resizebox{.85\hsize}{!}{$
		VASR(A^{adv}, B^{adv}; f) = ASR(A^{adv}; f) - ASR(B^{adv}; f).
		$}
\end{equation}
However, this metric ignores the impact of changes in correctly classified examples, which does not accurately reflect the performance of IEs. Therefore, we propose the immune rate (IR) to accurately measure the performance of IEs.

As shown in Table~\ref{tab:cf1} and~\ref{tab:cf2}, the IR for $B$ of $A$ against $f$ and $g$ is defined as
\begin{equation}
	\begin{aligned}
		IR(B; A, f, g) = \frac{\left | A_2 \cap B_1 \right |}{\left | A_2 \cap (B_1 \cup B_2) \right |}.
	\end{aligned}
\end{equation}
The IR excludes the raw images and IEs that were misclassified, as well as the invalid adversarial examples at first. This metric accurately reflects the percentage of the invalidation of adversarial perturbations due to IPs. The IR ranges from $0$ to $1$. When $B_1 = \varnothing$, the IR is $0$, indicating the worst performance of the IEs. When $B_2 = \varnothing$, the IR is $1$, indicating the best performance of the IEs. Note that $A_1$, $A_2$, $B_1$ and $B_2$ exclude the raw image whose ground-truth label is the target label for the ASR, VASR and IR of targeted attacks.

\section{Experiments and analysis}

\subsection{Experimental settings} \label{sec:exp_setting}

Since generation-based attacks not only have strong attack capabilities but also exhibit fast generation speed, which is not typically found in other types of attacks~\cite{xiaogenerating}, we craft IEs for generation-based attacks to defend against them in our experiments. Specifically, we choose five distinct generation-based attacks, \ie, ATN$_\text{0}$~\cite{baluja2018learning} (the target label is $0$), UAN~\cite{hayes2018learning}, AdvGAN~\cite{xiaogenerating}, AdvGAN++~\cite{jandial2019advgan++}, and RGAN~\cite{zhang2022self}. These attacks cover a range of techniques, including targeted and non-targeted attacks, universal and non-universal perturbations, and various network architectures (\eg, autoencoder, decoder, and GAN). According to Assumption~\ref{list:one}, we configure all generation-based attacks to white-box attacks to achieve the best attack performance and set the hyperparameters to official values provided in corresponding papers. We craft IEs for advanced GAN-based attacks, i.e., AdvGAN and AdvGAN++. We evaluate the white-box performance of the IEs against the source attacks and evaluate their transferability against the target attacks.

\begin{table}[t]
	\begin{center}
		\resizebox{0.45\textwidth}{!}{%
			\begin{tabular}{|l|ccccc|}
				\hline
				Datasets & ATN$_\text{0}$ & UAN  & AdvGAN & AdvGAN++ & RGAN \\ \hline\hline
				CIFAR-10                         & 98.0 & 85.9 & 96.4   & 98.6     & 96.4 \\ \hline
				MNIST                           & 78.9 & 78.1 & 97.9   & 98.0     & 99.5 \\ \hline
			\end{tabular}%
		}
	\end{center}
	\caption{The ASRs of various generation-based attacks on different datasets.}
	\label{tab:asr}
\end{table}
Since above generation-based attacks all involve small-size datasets, we choose CIFAR-10~\cite{krizhevsky2009learning} and MNIST~\cite{lecun1998gradient} for our experiments. For the classifiers, we select Inc-v3-like~\cite{szegedy2016rethinking} with an accuracy of $92.4\%$ for CIFAR-10, and a CNN~\cite{lecun1989backpropagation} with an accuracy of $99.1\%$ for MNIST. For the generation-based attacks, we report the ASRs on different datasets in Table~\ref{tab:asr}.

\begin{table}[t]
	\begin{center}
		\begin{tabular}{@{}|l|cc|cc|@{}}
			\hline
			\multirow{2}{*}{Parameters} & \multicolumn{2}{c|}{CIFAR-10} & \multicolumn{2}{c|}{MNIST} \\ \cline{2-5}
			& Opt.    & Grad.    & Opt.   & Grad.   \\ \hline\hline
			$\lambda$                              & 0.1          & 0.1          & 0.1         & 0.1         \\
			$\eta$                              & 100.0        & 0.0          & 10.0        & 0.0         \\
			$\tau$                              & 16           & 32           & 32          & 64          \\
			$T$                                & 500          & 5            & 1000        & 5           \\
			lr/$\alpha$                         & $10^{-3}$         &   24           & $10^{-3}$        &     48        \\
			$L_p$                                & $L_{\infty}$          & $L_{\infty}$          & $L_{\infty}$         & $L_{\infty}$         \\ 
			\hline
		\end{tabular}
	\end{center}
	\caption{Hyperparameters for different datasets and immune defenses. $T$ denotes the number of iterations. ``lr" denotes the learning rate for Adam. ``$L_p$" indicates that the IPs are subjected to the constraint of $L_{\infty}$ norm.}
	\label{tab:params}
\end{table}
In the white-box immune defense, We evaluate the performance of gradient-based (Grad.) and optimization-based (Opt.) approaches. In the black-box immune defense, we compare the transferability of IEs crafted by GSD and MGSD. Furthermore, we tranfer Polyak Momentum (PM)~\cite{polyak1964some,dong2018boosting} and Variance Tuning (VT)~\cite{wang2021enhancing}, which demonstrate effectiveness in improving the transferability of adversarial examples, into immune defense and integrate them into MGSD, denoted as PM-MGSD and VT-MGSD, to further validate the effectiveness of the proposed method. The hyperparameters in our experiments are shown in Table~\ref{tab:params}.

\subsection{Comparison for white-box immune denfenses} \label{sec:wid}

\begin{table*}[t]
	\begin{center}
		\resizebox{0.95\textwidth}{!}{%
			\begin{tabular}{|l|c|c|ccccccc|}
				\hline
				Datasets &
				Attacks &
				Defenses &
				Accuracy &
				ATN$_{\text{0}}$ &
				UAN &
				AdvGAN &
				AdvGAN++ &
				RGAN &
				$L_{\infty}$ \\ \hline\hline
				\multirow{4}{*}{CIFAR-10} &
				\multirow{2}{*}{AdvGAN} &
				Grad. &
				90.0 &
				\textbf{41.7/39.7} &
				\textbf{67.3/61.1} &
				76.4/79.3* &
				\textbf{16.7/17.0} &
				\textbf{71.1/73.8} &
				32 \\
				&
				&
				Opt. &
				\textbf{100.0} &
				13.0/11.9 &
				23.8/22.5&
				\textbf{96.4/100.0*} &
				-2.6/0.7 &
				33.9/28.6 &
				\textbf{11} \\ \cline{2-10} 
				&
				\multirow{2}{*}{AdvGAN++} &
				Grad. &
				81.9 &
				\textbf{43.7/40.2} &
				\textbf{69.8/68.9} &
				\textbf{19.1/19.8} &
				84.7/85.8* &
				\textbf{23.1/23.8} &
				32 \\ 
				&
				&
				Opt. &
				\textbf{100.0} &
				13.9/12.4 &
				24.2/23.3 &
				3.2/5.6 &
				\textbf{98.6/100.0*} &
				8.6/6.4 &
				\textbf{8} \\ \hline
				\multirow{4}{*}{MNIST} &
				\multirow{2}{*}{AdvGAN} &
				Grad. &
				99.4 &
				\textbf{42.5/29.7} &
				\textbf{8.8/14.7} &
				94.4/96.4* &
				\textbf{35.2/36.5} &
				\textbf{78.5/78.7} &
				64 \\ 
				&
				&
				Opt. &
				\textbf{100.0} &
				0.9/0.0 &
				4.5/11.0 &
				\textbf{96.3/98.4*} &
				1.9/3.9 &
				42.5/41.1 &
				\textbf{26} \\ \cline{2-10} 
				&
				\multirow{2}{*}{AdvGAN++} &
				Grad. &
				86.5 &
				\textbf{42.6/29.5} &
				\textbf{-2.8/3.7} &
				\textbf{18.8/20.5} &
				88.0/89.8* &
				\textbf{25.8/25.8} &
				64 \\ 
				&
				&
				Opt. &
				\textbf{98.4} &
				5.4/5.0 &
				-7.6/0.9 &
				2.7/3.3 &
				\textbf{96.4/98.4*} &
				6.6/6.6 &
				\textbf{30} \\ \hline
			\end{tabular}%
		}
	\end{center}
	\caption{The accuracy (\%), VASRs (\%), IRs (\%), and $L_{\infty}$ norm of two white-box immune defenses against the white-box and black-box attacks, respectively. Data in ``VASR/IR" format represents the VASR and IR, respectively. For instance, in the cell ``41.7/39.7," 41.7 indicates the VASR as 41.7\%, while 39.7 indicates the IR as 39.7\%. The symbol ``*" indicates the white-box immune defense.}
	\label{tab:wid}
\end{table*}
We evaluate the performance of two white-box immune defenses, \ie, Grad. (Eq.~\ref{eq:ie_optim} + GSD) and Opt. (Eq.~\ref{eq:ie_opt1} + Adam), respectively. The experimental setup was the same as described in Sec.~\ref{sec:exp_setting}. Table~\ref{tab:wid} summarizes the accuracy, VASRs, and IRs of the IEs, as well as the $L_{\infty}$ norm of the IPs. It is noteworthy that the accuracy, VASRs, IRs, and $L_{\infty}$ norm of the IEs crafted by Opt. is superior to Grad. in the white-box immune defense. Therefore, we recommend the optimization-based approach in the white-box immune defense. However, Opt. is less transferable than Grad. in the black-box immune defense. The results demonstrate that the hypothesis in Sec.~\ref{sec:gmm} is right, \ie, gradient-based immune defenses have higher transferability than optimization-based immune defenses but lower performance in the white-box immune defense.

\subsection{Comparison for transferability}

\begin{table*}[t]
	\begin{center}
		\resizebox{0.95\textwidth}{!}{%
			\begin{tabular}{|l|c|c|cccccc|}
				\hline
				Datasets &
				Attacks &
				Defenses &
				Accuracy &
				ATN$_{\text{0}}$ &
				UAN &
				AdvGAN &
				AdvGAN++ &
				RGAN \\ \hline\hline
				\multirow{12}{*}{CIFAR-10} &
				\multirow{6}{*}{AdvGAN} &
				GSD &
				90.0 &
				41.7/39.7 &
				67.3/61.1 &
				76.4/79.3* &
				16.7/17.0 &
				71.1/73.8 \\
				&
				&
				MGSD (\textbf{Ours}) &
				\textbf{93.7} &
				\textbf{46.3/44.3} &
				\textbf{71.1/68.3} &
				\textbf{79.8/83.3*} &
				\textbf{17.9/18.5} &
				\textbf{75.4/78.7} \\ \cline{3-9} 
				&
				&
				PM-GSD &
				97.3 &
				71.2/67.7 &
				80.9/89.3 &
				94.2/97.7* &
				32.1/32.6 &
				91.7/95.1 \\
				&
				&
				PM-MGSD (\textbf{Ours}) &
				\textbf{97.5} &
				\textbf{73.1/70.8} &
				\textbf{81.3/90.2} &
				\textbf{94.3/98.0*} &
				\textbf{32.9/34.3} &
				\textbf{92.2/95.8} \\ \cline{3-9} 
				&
				&
				VT-GSD &
				98.2 &
				79.9/74.9 &
				82.4/92.4 &
				95.0/98.4* &
				32.9/33.3 &
				92.6/95.8 \\
				&
				&
				VT-MGSD (\textbf{Ours}) &
				\textbf{99.0} &
				\textbf{83.7/78.5} &
				\textbf{83.2/93.7} &
				\textbf{96.0/99.5*} &
				\textbf{35.5/36.0} &
				\textbf{93.9/96.3} \\ \cline{2-9} 
				&
				\multirow{6}{*}{AdvGAN++} &
				GSD &
				81.9 &
				43.7/40.2 &
				69.8/68.9 &
				19.1/19.8 &
				84.7/85.8* &
				23.1/23.8 \\
				&
				&
				MGSD (\textbf{Ours}) &
				\textbf{95.4} &
				\textbf{45.6/41.8} &
				\textbf{71.2/69.7} &
				\textbf{19.8/21.3} &
				\textbf{92.9/94.5*} &
				\textbf{24.4/26.1} \\ \cline{3-9} 
				&
				&
				PM-GSD &
				97.8 &
				72.5/71.0 &
				83.0/93.9 &
				39.6/41.2 &
				98.2/99.6* &
				47.1/49.0 \\
				&
				&
				PM-MGSD (\textbf{Ours}) &
				\textbf{98.9} &
				\textbf{73.5/74.6} &
				\textbf{83.8/95.6} &
				\textbf{41.7/44.8} &
				\textbf{98.4/99.8*} &
				\textbf{49.6/53.0} \\ \cline{3-9} 
				&
				&
				VT-GSD &
				98.5 &
				76.6/74.1 &
				82.8/92.7 &
				40.3/41.8 &
				98.4/98.8* &
				51.7/53.4 \\
				&
				&
				VT-MGSD (\textbf{Ours}) &
				\textbf{99.2} &
				\textbf{81.2/79.3} &
				\textbf{84.7/96.6} &
				\textbf{46.7/48.4} &
				\textbf{98.6/100.0*} &
				\textbf{57.5/59.4} \\ \hline
				\multirow{12}{*}{MNIST} &
				\multirow{6}{*}{AdvGAN} &
				GSD &
				99.4 &
				42.5/29.7 &
				8.8/14.7 &
				\textbf{94.4/96.4*} &
				35.2/36.5 &
				78.5/78.7 \\
				&
				&
				MGSD (\textbf{Ours}) &
				\textbf{99.8} &
				\textbf{46.9/34.9} &
				\textbf{31.9/41.8} &
				91.3/93.4* &
				\textbf{49.5/50.8} &
				\textbf{90.2/90.6} \\ \cline{3-9} 
				&
				&
				PM-GSD &
				\textbf{100.0} &
				45.1/32.8 &
				20.9/27.7 &
				\textbf{97.9/100.0*} &
				37.2/38.2 &
				97.5/98.0 \\
				&
				&
				PM-MGSD (\textbf{Ours}) &
				\textbf{100.0} &
				\textbf{48.2/38.1} &
				\textbf{38.9/50.4} &
				\textbf{97.9/100.0*} &
				\textbf{59.2/60.5} &
				\textbf{98.2/98.7} \\ \cline{3-9} 
				&
				&
				VT-GSD &
				\textbf{100.0} &
				46.8/35.8 &
				37.4/48.3 &
				\textbf{97.9/100.0*} &
				54.8/56.3 &
				97.2/97.7 \\
				&
				&
				VT-MGSD (\textbf{Ours}) &
				\textbf{100.0} &
				\textbf{49.7/40.5} &
				\textbf{49.2/63.2} &
				\textbf{97.9/100.0*} &
				\textbf{65.8/67.3} &
				\textbf{98.3/98.8} \\ \cline{2-9} 
				&
				\multirow{6}{*}{AdvGAN++} &
				GSD &
				86.5 &
				42.6/29.5 &
				-2.8/3.7 &
				18.8/20.5 &
				88.0/89.8* &
				25.8/25.8 \\
				&
				&
				MGSD (\textbf{Ours}) &
				\textbf{99.8} &
				\textbf{45.4/32.2} &
				\textbf{14.9/22.4} &
				\textbf{59.9/61.6} &
				\textbf{93.4/95.3*} &
				\textbf{69.5/69.7} \\ \cline{3-9} 
				&
				&
				PM-GSD &
				97.3 &
				45.3/30.7 &
				-2.4/4.8 &
				39.5/41.1 &
				\textbf{97.9/99.9*} &
				48.7/48.7 \\ 
				&
				&
				PM-MGSD (\textbf{Ours}) &
				\textbf{100.0} &
				\textbf{46.0/33.5} &
				\textbf{16.7/24.6} &
				\textbf{64.0/65.7} &
				\textbf{97.9/99.9*} &
				\textbf{74.8/75.0} \\ \cline{3-9} 
				&
				&
				VT-GSD &
				97.5 &
				46.9/32.6 &
				6.9/15.3 &
				50.1/51.7 &
				\textbf{97.7/99.7*} &
				61.9/62.0 \\
				&
				&
				VT-MGSD (\textbf{Ours}) &
				\textbf{100.0} &
				\textbf{47.9/36.9} &
				\textbf{21.2/30.3} &
				\textbf{72.7/74.5} &
				\textbf{97.7/99.7*} &
				\textbf{84.5/84.9} \\ \hline
			\end{tabular}%
		}
	\end{center}
	\caption{The VASRs (\%), IRs (\%), and Accuracy (\%) of various immune defenses against adversarial attacks. The IEs are crafted for AdvGAN and AdvGAN++, respectively. PM-GSD and VT-GSD indicate the integration of PM~\cite{polyak1964some,dong2018boosting} and VT~\cite{wang2021enhancing} with GSD, respectively. The symbol ``*" indicates the white-box immune defense.}
	\label{tab:bid}
\end{table*}

In the black-box immune defense, we focus on the gradient-based approach with relatively high basic transferability. We compare the transferability of IEs crafted by vanilla GSD and the proposed MGSD. Furthermore, We integrate PM and VT into the proposed method to further improve the transferability of IEs. We present the accuracy, VASRs, and IRs of black-box immune defenses in Table~\ref{tab:bid}, where the IEs are crafted for AdvGAN and AdvGAN++ on CIFAR-10 and MNIST, respectively.

The experimental results demonstrate that MGSD not only improves the accuracy of the examples but also significantly improves the transferability of IEs more than baselines. In addition, PM and VT transfer well into immune defense and further improve the performance of MGSD. Particularly, our best method,~\ie, VT-MGSD, achieve an accuracy of $99.0\%\sim100.0\%$, an average VASR of $66.0\%$, and an average IR of $67.8\%$ against the black-box generation-based attacks, respectively.

\subsection{Ablation study}

In this paper, ablation studies are only conducted on black-box immune defenses. Without loss of generality, we utilize our best method,~\ie, VT-MGSD, to craft IEs for AdvGAN against target attcks on CIFAR-10 and MNIST, respectively. We conduct an investigation how the values of the number of iterations $T$, the size of immune perturbation $\tau$, and the step size $\alpha$ affect the transferability of IEs.

\subsubsection{The number of iterations}

\begin{figure}[t]
	\begin{center}
		\begin{minipage}[b]{.48\linewidth}
			\begin{center}
				\includegraphics[width=1.0\linewidth]{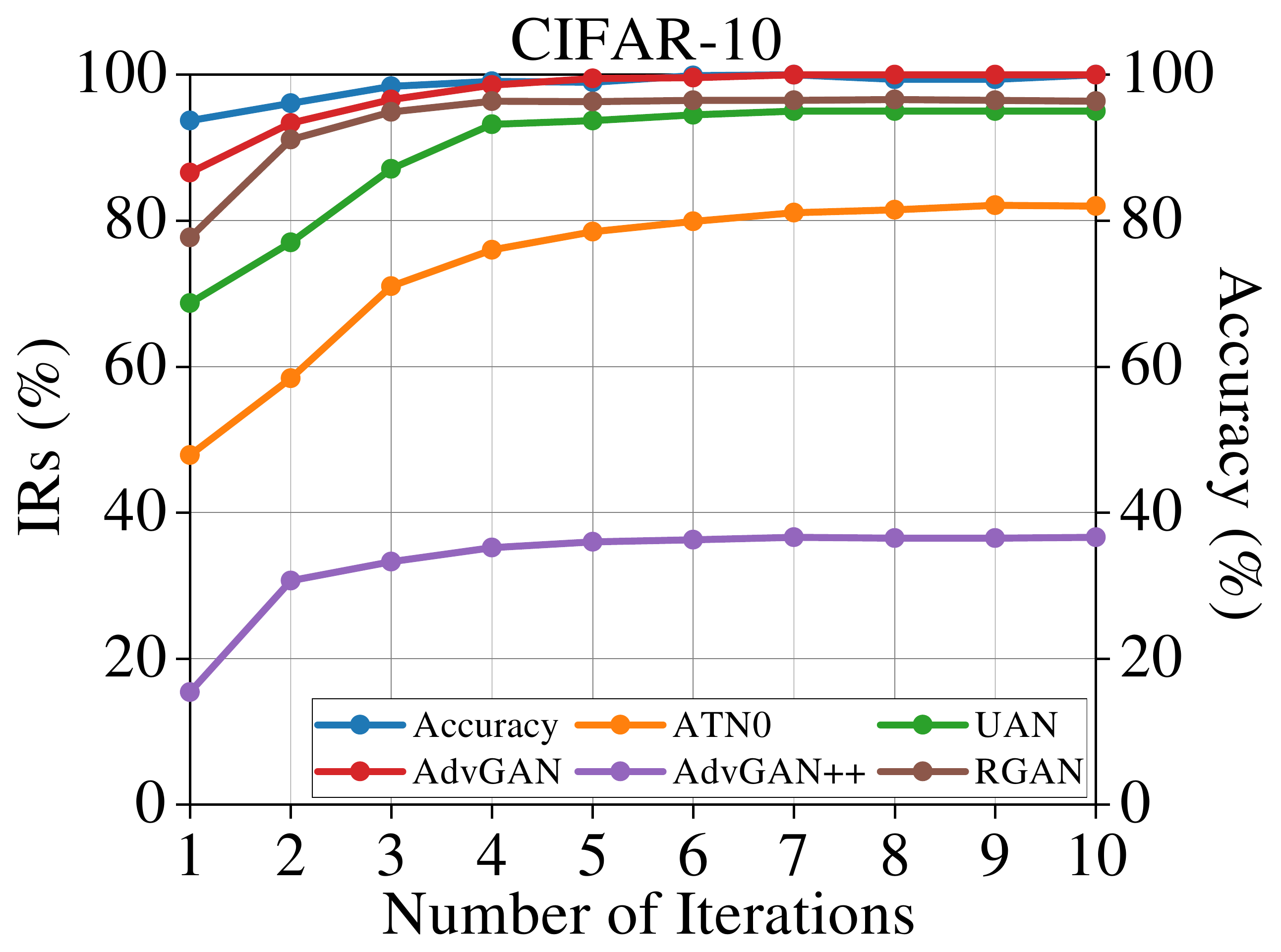}
			\end{center}
		\end{minipage}
		\begin{minipage}[b]{.48\linewidth}
			\begin{center}
				\includegraphics[width=1.0\linewidth]{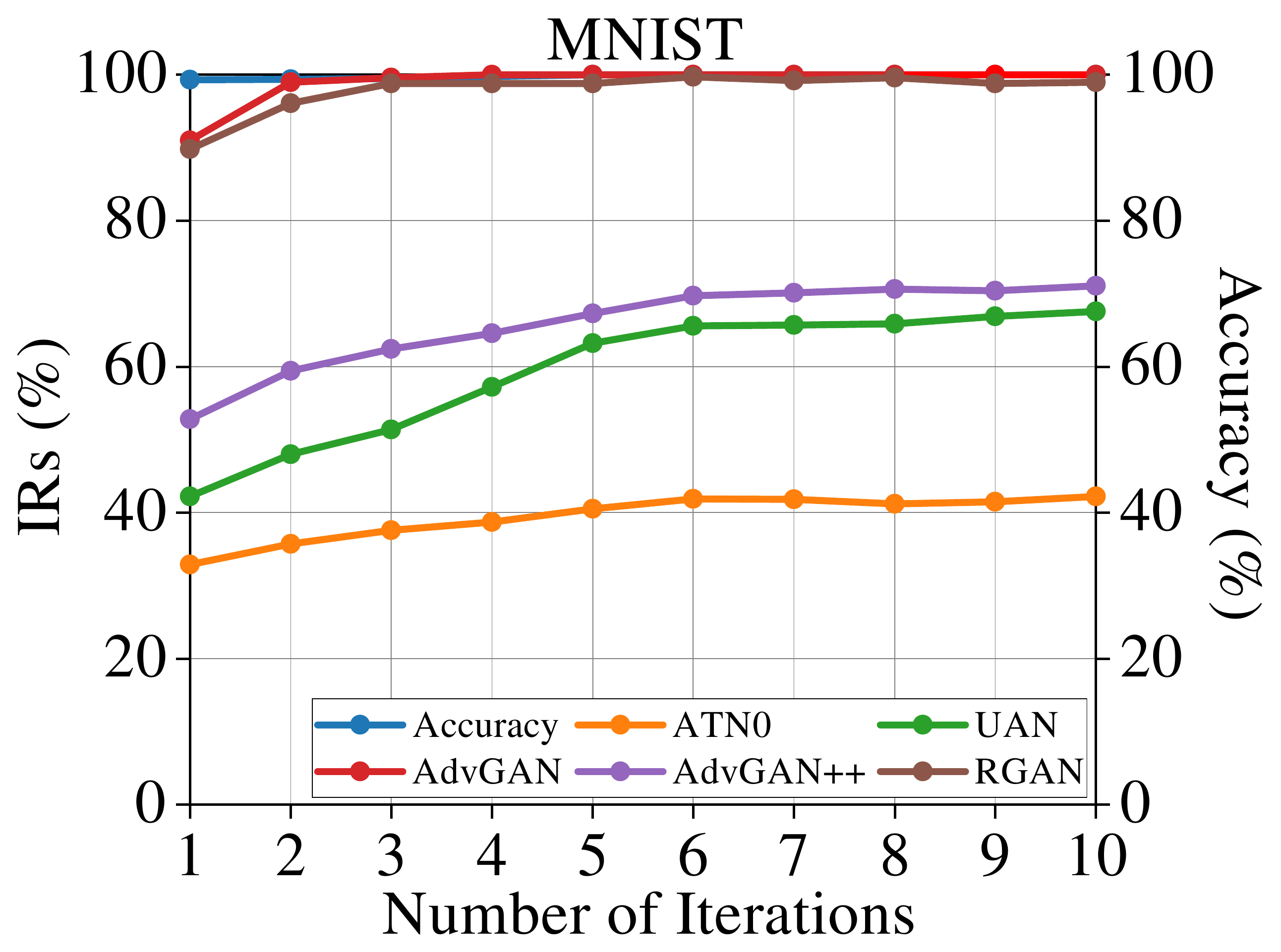}
			\end{center}
		\end{minipage}
	\end{center}
	\caption{The accuracy (\%) and IRs (\%) of the IEs crafted by VT-MGSD for AdvGAN. The IEs defend against AdvGAN (white-box immune defense), ATN$_\text{0}$, UAN, AdvGAN++ and RGAN (black-box immune defense), with the number of iterations ranging from $1$ to $10$.}
	\label{fig:num_of_t}
\end{figure}

To investigate the effect of the number of iterations $T$ on the transferability of IEs, we pre-set $(\tau, \alpha) = (32, 24)$ for CIFAR-10 and $(\tau, \alpha) = (64, 48)$ for MNIST, and vary $T$ from $1$ to $10$ with a step size of $1$. We then evaluate the accuracy and IRs of the IEs against black-box attcks. The results are shown in Fig.~\ref{fig:num_of_t}. The table reports that, as the number of iterations $T$ increases, the IRs against the black-box attcks increase and gradually converge. Additionally, the accuracy of IEs and the IRs against the source attcks nearly reach $100\%$. However, as $T$ increases, the computational cost also increases. Thus, we set $T = 5$ to balance the computational cost and the transferability of IEs.

\subsubsection{The size of immune perturbation}

\begin{figure}[t]
	\begin{center}
		\begin{minipage}[b]{.48\linewidth}
			\begin{center}
				\includegraphics[width=1.0\linewidth]{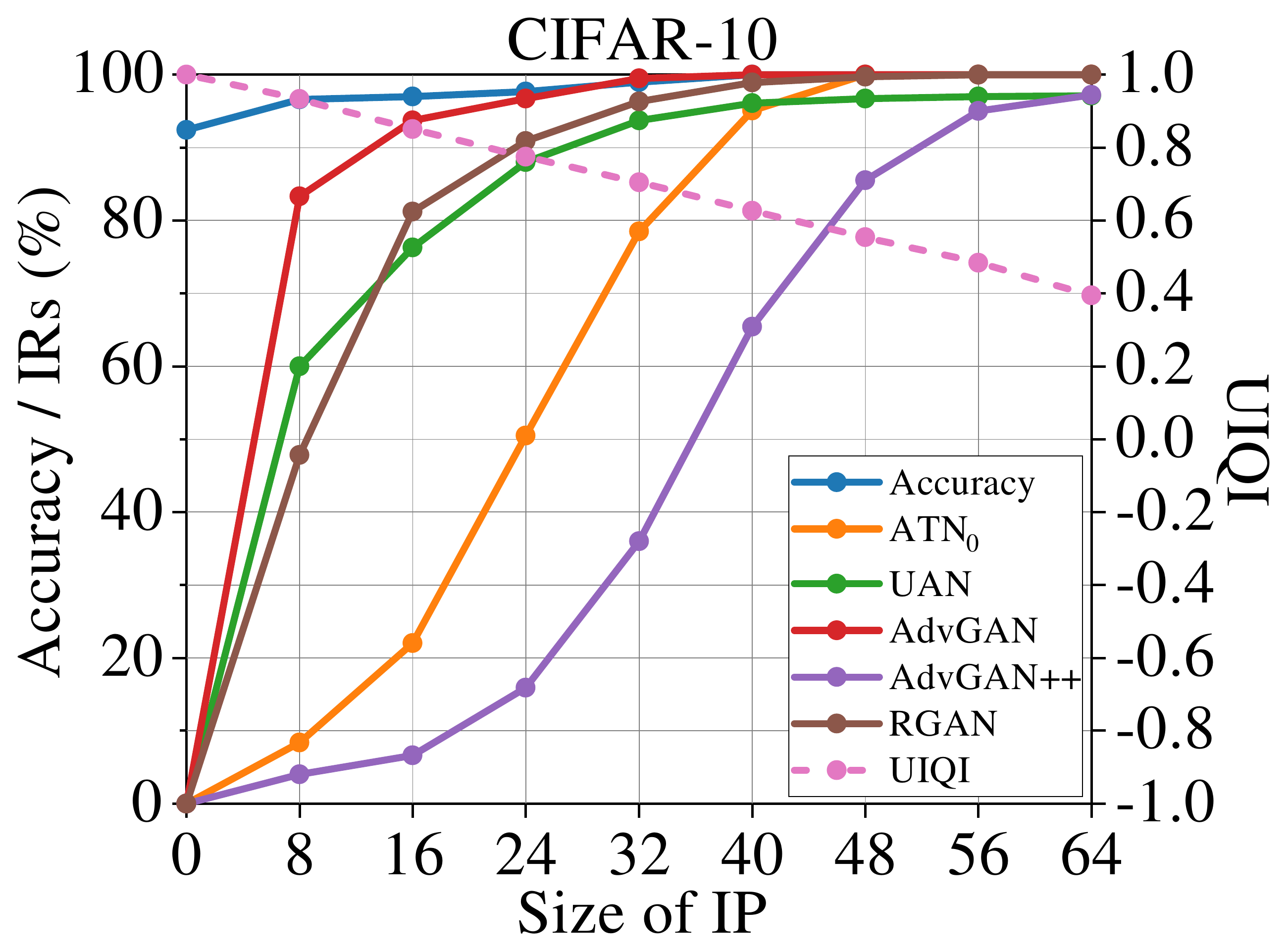}
			\end{center}
		\end{minipage}
		\begin{minipage}[b]{.48\linewidth}
			\begin{center}
				\includegraphics[width=1.0\linewidth]{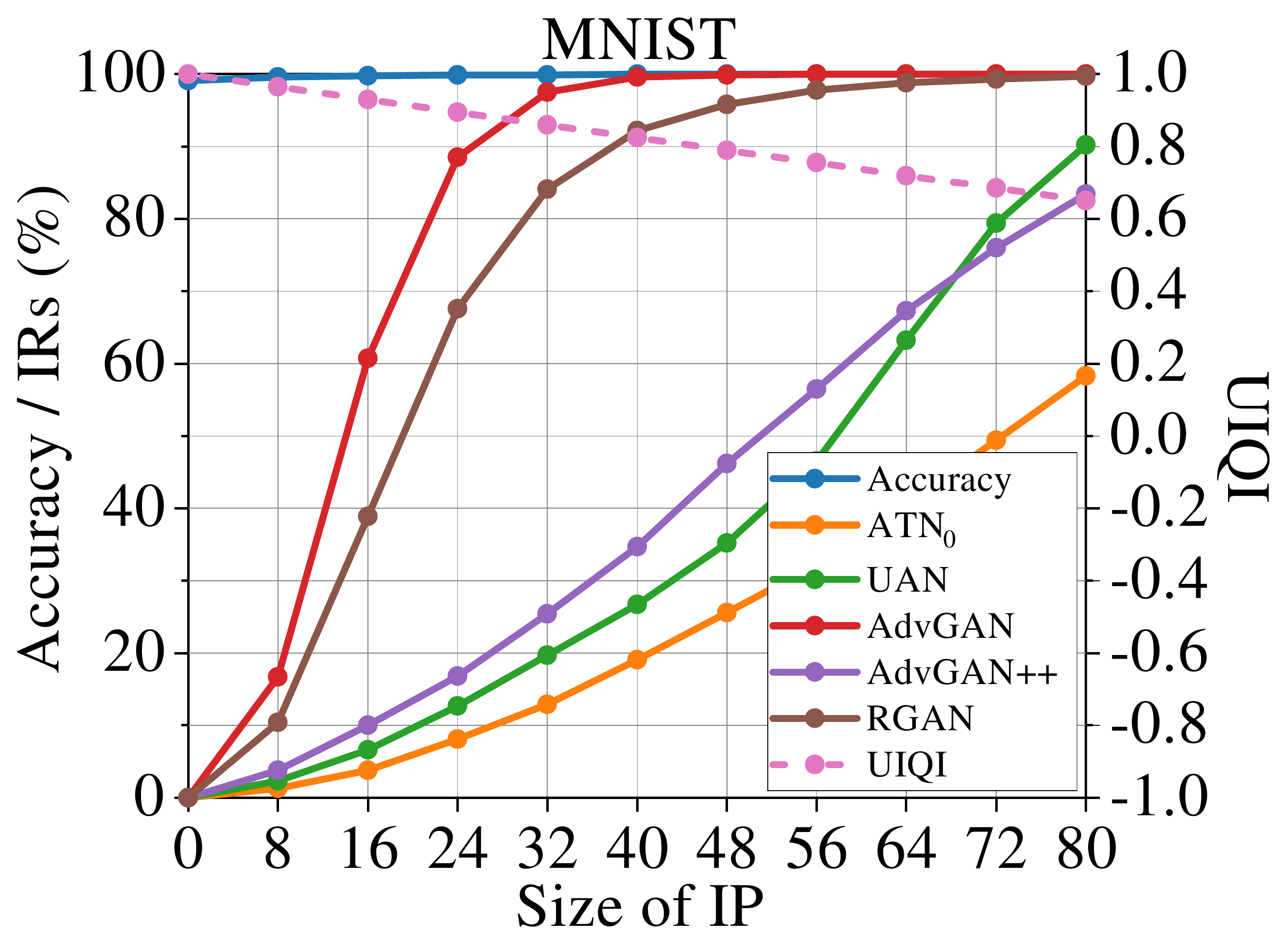}
			\end{center}
		\end{minipage}
	\end{center}
	\caption{The accuracy (\%), IRs (\%) and UIQI of the IEs crafted by VT-MGSD for AdvGAN. The IEs defend against AdvGAN (white-box immune defense), ATN$_\text{0}$, UAN, AdvGAN++ and RGAN (black-box immune defense), with various sizes of immune perturbation.}
	\label{fig:max_ip}
\end{figure}

In this experiment, we craft IEs for AdvGAN and investigate the impact of the size of immune perturbation $\tau$ on the transferability of IEs. For CIFAR-10, we pre-set $(T, \alpha) = (5, 24)$ and vary $\tau$ from $0$ to $64$ with a step size of $8$, while for MNIST, we pre-set $(T, \alpha) = (5, 48)$ and vary $\tau$ from $0$ to $80$ with a step size of $8$. The accuracy, IRs, and UIQI\footnote{The Universal Image Quality Index (UIQI)~\cite{wang2002universal} is a visual quality metric that ranges from $-1$ to $1$, where higher values correspond to better visual quality. Compared to other image quality assessment metrics (\eg, PSNR~\cite{almohammad2010stego} and SSIM~\cite{wang2004image}), UIQI is more sensitive to changes in brightness, contrast, and color, thus providing better robustness and accuracy.} of IEs are evaluated to explore the effect of different values of $\tau$ on the transferability of IEs. We report the experimental results in Fig.~\ref{fig:max_ip}. The accuracy and IRs of IEs increase as $\tau$ increases, while the UIQI decreases.To maintain the visual quality of the IEs, we selected the size of value of $\tau$ that satisfies $\text{UIQI} \ge 0.7$, thereby striking a balance between the transferability and visual quality of the IEs. Specifically, for CIFAR-10, we set $\tau = 32$ with a corresponding UIQI value of $0.705$, and for MNIST, we set $\tau = 64$ with a corresponding UIQI value of $0.719$.

\subsubsection{The step size}

\begin{figure}[t]
	\begin{center}
		\begin{minipage}[b]{.48\linewidth}
			\begin{center}
				\includegraphics[width=1.0\linewidth]{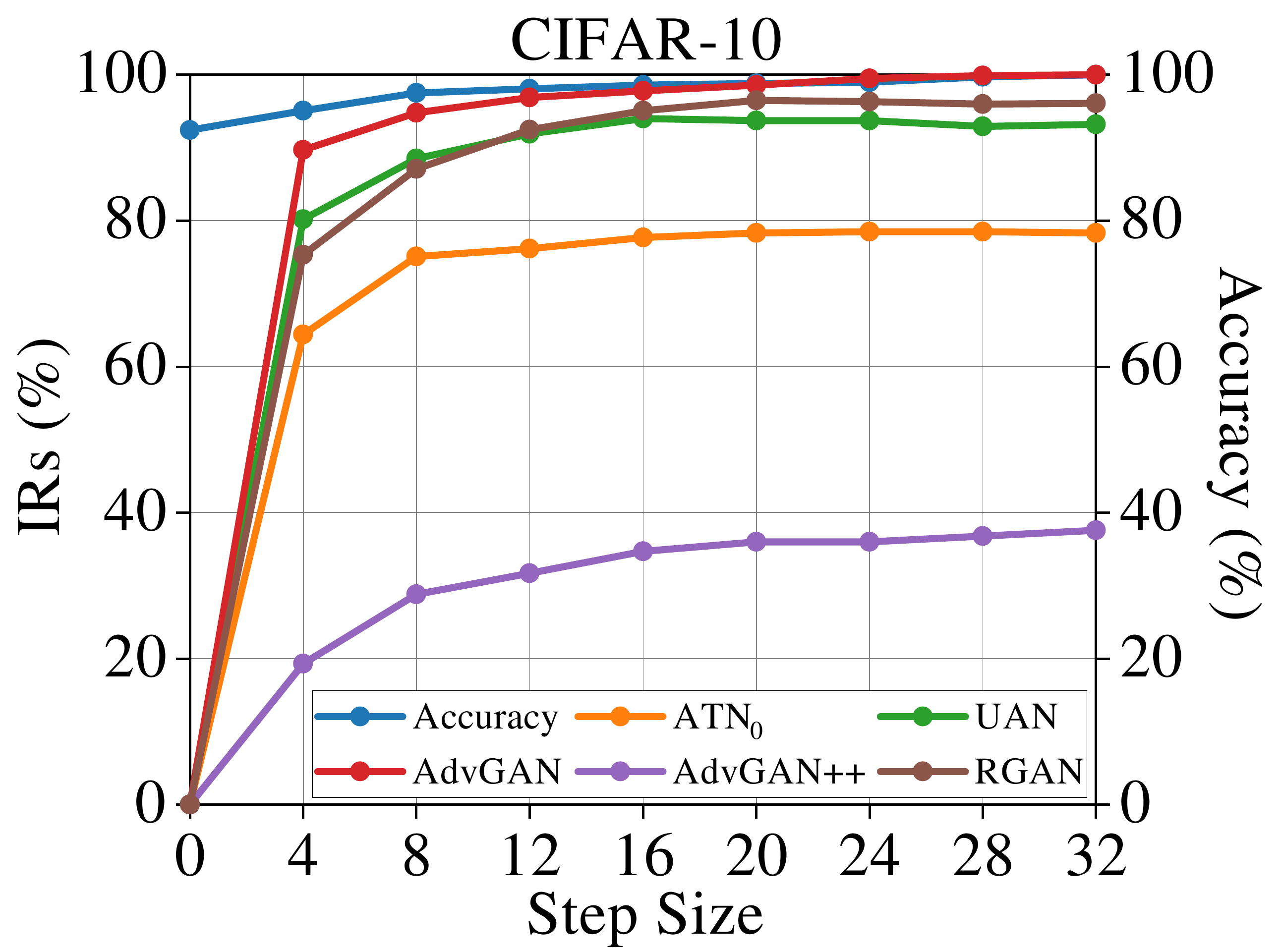}
			\end{center}
		\end{minipage}
		\begin{minipage}[b]{.48\linewidth}
			\begin{center}
				\includegraphics[width=1.0\linewidth]{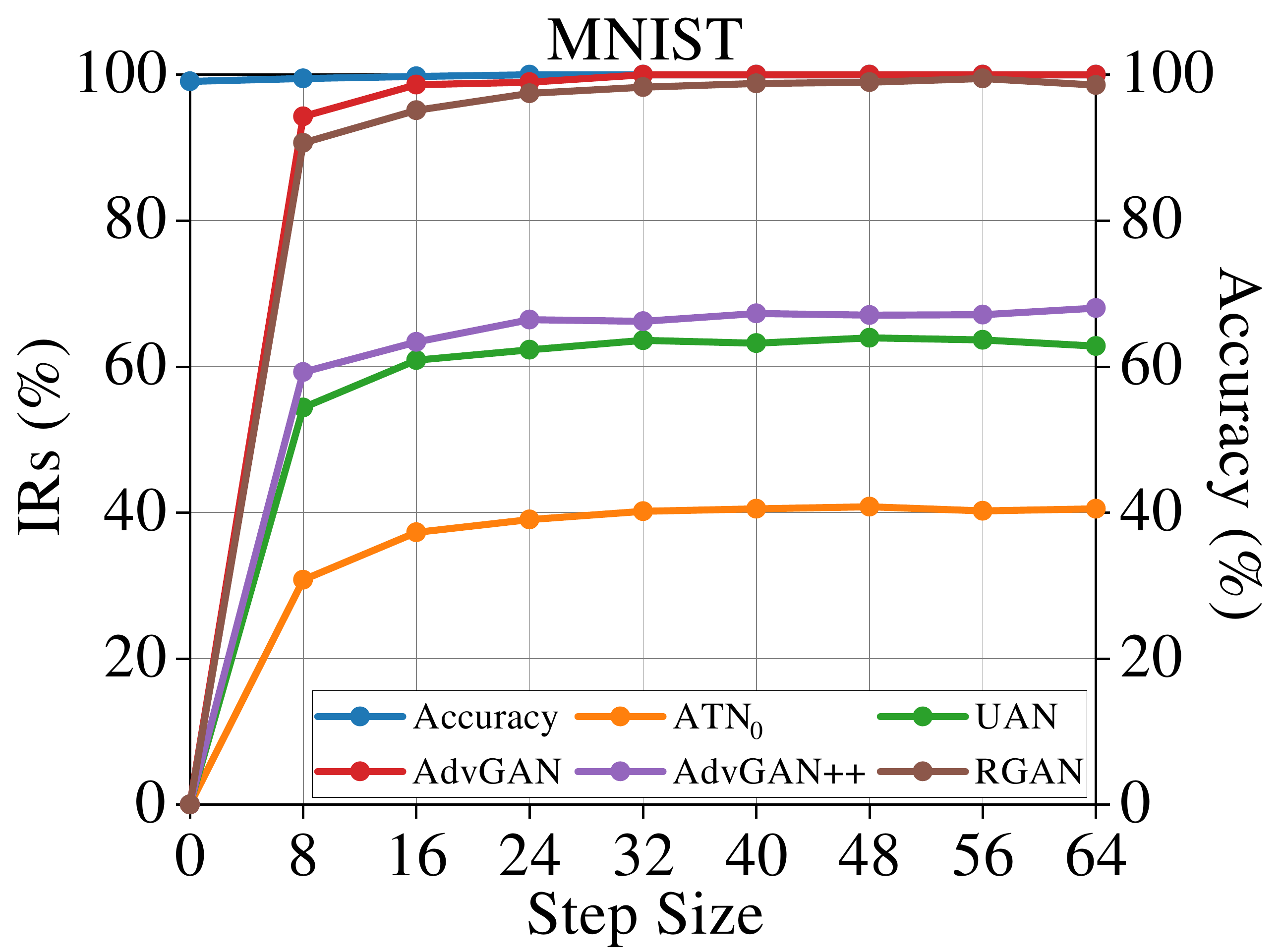}
			\end{center}
		\end{minipage}
	\end{center}
	\caption{The accuracy (\%) and IRs (\%) of the IEs crafted by VT-MGSD for AdvGAN. The IEs defend against AdvGAN (white-box immune defense), ATN$_\text{0}$, UAN, AdvGAN++ and RGAN (black-box immune defense), with various step sizes.}
	\label{fig:step_size}
\end{figure}

We finally explore the effect of a on the transferability of IEs. For CIFAR-10, we pre-set $(T, \tau) = (5, 32)$ and vary $\alpha$ from $0$ to $32$ with a step size of $4$, while for MNIST, we pre-set $(T, \tau) = (5, 64)$ and vary $\alpha$ from $0$ to $64$ with a step size of $8$. The accuracy and IRs of IEs against source attcks and four black-box target attcks are illustrated in Fig.~\ref{fig:step_size}. It can be observed that the accuracy and IRs of IEs improve and gradually converge with increasing step size $\alpha$. This phenomenon can be attributed to the higher likelihood of overfitting the IE to the source attck with smaller step sizes, resulting in poor transferability under the same size of immune perturbation and number of iterations. Eventually, we set $\alpha = 24$ for CIFAR-10 and $\alpha = 48$ to ensure stable performance.

\section{Conclusion}

In this work, we propose a novel adversarial defense mechanism to prevent the generation of adversarial examples. Specifically, we first introduce the concept of Immune Examples (IEs) and immune defense. Then, We propose a gradient-based and an optimization-based approach, respectively, for crafting IEs in the white-box immune defense. Additionally, we also explore the black-box immune defense and propose the Masked Gradient Sign Descent (MGSD) to improve the transferability of IEs. The experimental results demonstrate that the optimization-based approach can generate IEs with high performance and high visual quality in the white-box immune defense. The gradient-based approach can generate immune examples with high transferability in the black-box immune defense, and the proposed MGSD can further improve the transferability. This work is expected to further improve the security of examples and DNNs. However, since immune perturbations may be detected and disrupted by attackers, we will focus on improving the robustness of IEs in future work.

{\small
	\bibliographystyle{plain}
	\bibliography{references}
}

\clearpage
\appendix

\section{Appendix}

\subsection{Proofs and analysis of formulas}

\subsubsection{Proofs and analysis of Eq.~\ref{eq:adv_delta} and Eq.~\ref{eq:error_limit}}

According to Eq.~\ref{eq:ie_update2} and ignoring the truncation, the variation of adversarial examples in the $k$-th dimension, \ie, $\Delta_k^t$, can be rewritten as follows:
\begin{align}
	\Delta_k^t
	&= g(\x_{t+1}^{IE})_k - g(\x_t^{IE})_k \notag\\
	&= g(\x_t^{IE} + \len)_k - g(\x_t^{IE})_k, \label{eq:variation}
\end{align}
where $\len = - \alpha \cdot \mathrm{sign}(\nabla_{\x_t^{IE}} \J(g(\x_t^{IE}), y))$. Since $\left \| \len \right \|_2 \ne 0$,
\begin{align}
	\Delta_k^t = \frac{g(\x_t^{IE} + \len)_k - g(\x_t^{IE})_k}{\left \| \len \right \|_2} \cdot \left \| \len \right \|_2.
\end{align}

However, we can assume $\left \| \len \right \|_2 \to 0$ and denote the approximation error as $e(\left \| \len \right \|_2)$, so that
\begin{align}
	\resizebox{.87\hsize}{!}{$
		\Delta_k^t = \left(\lim_{\left \| \len \right \|_2 \to 0} \frac{g(\x_t^{IE} + \len)_k - g(\x_t^{IE})_k}{\left \| \len \right \|_2} + e(\left \| \len \right \|_2)\right) \cdot \left \| \len \right \|_2
		$}.
\end{align}
It is not difficult to find that as $\left \| \len \right \|_2$ approaches $0$, $e(\left \| \len \right \|_2)$ also approaches $0$, which corresponds to Eq.~\ref{eq:error_limit}.

We can further deduce from the definition of directional derivative that
\begin{align}
	\Delta_k^t = \left( \nabla_{\len} g(\x_t^{IE})_k + e(\left \| \len \right \|_2) \right) \cdot \left \| \len \right \|_2.
\end{align}
The directional gradient has the following property:
\begin{align}
	\nabla_{\len} g(\x_t^{IE})_k = \left \langle \nabla_{\x_t^{IE}} g(\x_t^{IE})_k, \frac{\len}{\left \| \len \right \|_2} \right \rangle,
\end{align}
where $\left \langle \cdot, \cdot \right \rangle$ denotes inner product. Thus 
\begin{align}
	\Delta_k^t\!=\!\left( \left \langle \nabla_{\x_t^{IE}} g(\x_t^{IE})_k, \frac{\len}{\left \| \len \right \|_2} \right \rangle + e(\left \| \len \right \|_2) \right) \cdot \left \| \len \right \|_2.
\end{align}

Due to the chain rule
\begin{align}
	\resizebox{.87\hsize}{!}{$
		\nabla_{\x_t^{IE}} \J(g(\x_t^{IE}), y) = \nabla_{\x_t^{IE}} g(\x_t^{IE})_k \cdot \nabla_{g(\x_t^{IE})_k} \J(g(\x_t^{IE}), y)
		$},
\end{align}
we deduce that
\begin{align}
	\Delta_k^t\!&=\!\left \langle \frac{\nabla_{\x_t^{IE}} \J(g(\x_t^{IE}), y)}{\nabla_{g(\x_t^{IE})_k} \J(g(\x_t^{IE}), y)}, \len \right \rangle\!+\!\left \| \len \right \|_2\!\cdot\!e(\left \| \len \right \|_2) \notag \\
	&=\frac{\left \langle \nabla_{\x_t^{IE}} \J(g(\x_t^{IE}), y), \len \right \rangle}{\nabla_{g(\x_t^{IE})_k} \J(g(\x_t^{IE}), y)} + \left \| \len \right \|_2 \cdot e(\left \| \len \right \|_2).
	\label{eq:variation_form1}
\end{align}

Finally, we eliminate $\len$ in the inner product operation to derive that
\begin{align}
	\Delta_k^t = -\alpha\!\cdot\!\frac{\left \| \nabla_{\x_t^{IE}} \J(g(\x_t^{IE}), y) \right \|_1}{\nabla_{g(\x_t^{IE})_k} \J(g(\x_t^{IE}), y)}\!+\!\left \| \len \right \|_2\!\cdot\!e(\left \| \len \right \|_2).
\end{align}

In addition, we also provide a more concise proof for Eq.~\ref{eq:adv_delta}. Specifically, based on the Taylor formula of several variables, \ie,
\begin{align}
	\resizebox{.87\hsize}{!}{$
		g(\x_t^{IE} + \len)_k = g(\x_t^{IE})_k + \left \langle \nabla_{\x_t^{IE}} g(\x_t^{IE})_k, \len \right \rangle + O(\left \| \len \right \|_2)
		$},
\end{align}
where $O(\len)$ denotes an infinitesimal of higher order for $\left \| \len \right \|_2$, Eq.~\ref{eq:variation} can be rewritten as
\begin{align}
	\Delta_k^t = \left \langle \nabla_{\x_t^{IE}} g(\x_t^{IE})_k, \len \right \rangle + O(\left \| \len \right \|_2).
\end{align}

Similarly, applying the chain rule and eliminating $\len$ in the inner product operation, we can also deduce that
\begin{align}
	\Delta_k^t = -\alpha\!\cdot\!\frac{\left \| \nabla_{\x_t^{IE}} \J(g(\x_t^{IE}), y) \right \|_1}{\nabla_{g(\x_t^{IE})_k} \J(g(\x_t^{IE}), y)} + O(\left \| \len \right \|_2).
	\label{eq:variation_form2}
\end{align}

It is worth noting that, based on Eq.~\ref{eq:variation_form1} and Eq.~\ref{eq:variation_form2},
\begin{align}
	O(\left \| \len \right \|_2) = \left \| \len \right \|_2 \cdot e(\left \| \len \right \|_2).
\end{align}
Thus, we assume that $e(\left \| \len \right \|_2) \sim N(0, \sigma \cdot \left \| \len \right \|_2)$ is reasonable, which not only satisfies Eq.~\ref{eq:error_limit}, but also satisfies that $\left \| \len \right \|_2 \cdot e(\left \| \len \right \|_2)$ is an infinitesimal of higher order for $\left \| \len \right \|_2$.

\subsubsection{Proof and analysis of Eq.~\ref{eq:l_norm}}

\begin{align}
	\left \| \len \right \|_2 
	&= \sqrt{\sum_{k=1}^{n} \left( - \alpha \cdot \mathrm{sign}(\nabla_{{\x_t^{IE}}_k} \J(g(\x_t^{IE}), y)) \right)^2} \notag \\
	&= \alpha \cdot \sqrt{\sum_{k=1}^{n} \left | \mathrm{sign} \left(\nabla_{{\x_t^{IE}}_k} \J(g(\x_t^{IE}), y) \right) \right |} \notag \\
	&= \alpha \cdot \sqrt{\left \| \mathrm{sign} \left(\nabla_{\x_t^{IE}} \J(g(\x_t^{IE}), y) \right) \right \|_1} \notag \\
	&= \alpha \cdot \sqrt{\left \| \nabla_{\x_t^{IE}} \J(g(\x_t^{IE}), y) \right \|_0}.
	\label{eq:l_norm}
\end{align}

\subsubsection{Proof and analysis of Eq.~\ref{eq:mask}}

Let $\M$ be a mask, we mask $\nabla_{\x_t^{IE}} \J(g(\x_t^{IE}), y)$ with $\M$ and rewrite $\len$ as follows:
\begin{align}
	\len = - \alpha \cdot \mathrm{sign}(\M \odot \nabla_{\x_t^{IE}} \J(g(\x_t^{IE}), y)).
\end{align}
We expect, after masking, the update directions of $\x_t^{IE}$ and $g(\x_t^{IE})$ keep aligned to make the update of immune examples stable. In other words, we expect
\begin{gather}
	\left( \x_{t+1}^{IE} - \x_t^{IE} \right) \cdot \left( g(\x_{t+1}^{IE}) - g(\x_t^{IE}) \right) \notag \\
	\!=\!\frac{\len\!\cdot\!\left \langle \nabla_{\x_t^{IE}} \J(g(\x_t^{IE}), y), \len \right \rangle}{\nabla_{g(\x_t^{IE})} \J(g(\x_t^{IE}), y)}\!+\!\len\!\cdot\!\left \| \len \right \|_2\!\cdot\! e(\left \| \len \right \|_2)^n\!>\!0. \label{eq:ineq_exp}
\end{gather}

Since the error is reduced after the mask, we ignore the effect of the error on the inequality sign, so that we expect
\begin{gather}
	\frac{\len \cdot \left \langle \nabla_{\x_t^{IE}} \J(g(\x_t^{IE}), y), \len \right \rangle}{\nabla_{g(\x_t^{IE})} \J(g(\x_t^{IE}), y)} > 0,
\end{gather}
\ie,
\begin{gather}
	\frac{\M \odot \nabla_{\x_t^{IE}} \J(g(\x_t^{IE}), y)}{\nabla_{g(\x_t^{IE})} \J(g(\x_t^{IE}), y)} > 0,
\end{gather}
Thus we deduce Eq.~ref{eq:mask} so that the mask meets our expectations.

\end{document}